%% file: main.tex
\documentclass[conference]{IEEEtran}
\IEEEoverridecommandlockouts
\usepackage{amsmath,amssymb,amsfonts}
\usepackage[numbers]{natbib}
\usepackage[noend]{algorithmic}
\usepackage{graphicx}
\usepackage{textcomp}
\usepackage{enumitem}
\usepackage{hyperref}
\usepackage{balance} 
\usepackage{color,soul}
\usepackage{xcolor}
\usepackage{adjustbox}
\usepackage{float}
\usepackage{comment}
\usepackage{hhline}
\usepackage{caption}
\usepackage{subcaption}
\usepackage{amsmath}
\usepackage{tikz}
\usetikzlibrary{positioning}
\usepackage{pgfplots}
\usepackage{pdflscape}
\usepackage{pgfplotstable}
\pgfplotsset{compat=1.7}
\usepgfplotslibrary{groupplots}
\usepackage{multirow}
\usepackage{algorithm}

\usepackage{comment}
\usepackage{colortbl}

\def\BibTeX{{\rm B\kern-.05em{\sc i\kern-.025em b}\kern-.08em
    T\kern-.1667em\lower.7ex\hbox{E}\kern-.125emX}}

\makeatletter
\def\addlegendimage{\pgfplots@addlegendimage}
\makeatother

\title{Training Models to Detect Successive Robot Errors from Human Reactions}

\author{\IEEEauthorblockN{Shannon Liu}
\IEEEauthorblockA{\textit{Cornell University} 
}
\and
\IEEEauthorblockN{Maria Teresa Parreira}
\IEEEauthorblockA{\textit{Cornell Tech} 
}
\and
\IEEEauthorblockN{Wendy Ju}
\IEEEauthorblockA{\textit{Cornell Tech} 
}}

\begin{document}

\maketitle


\input{sections/original}
\input{sections/methods}
\input{sections/models}

\input{sections/results}
\input{sections/discussion}
\input{sections/conclusion}

\bibliographystyle{abbrvnat}
\balance
\bibliography{references.bib}


\end{document}

%% file: sections/original.tex
\section{Introduction}
\label{sec:introduction}

As robots are increasingly integrated into society to collaborate with humans, developing systems for effective detection of robot errors becomes crucial for efficient human-robot interactions (HRI). When a robot fails multiple times, how will it know to change its behavior? Interestingly, humans react to robot errors through various verbal and nonverbal communication methods. Recent work has demonstrated that over the course of successive robot failures, human reactions intensify over time: reactions begin with confusion and adjustments in verbal communication and escalate to visible frustration shown through raised eyebrows, pursed lips, speaking in impatient or demanding tones, or hesitation before speaking \cite{lbr}. While existing work shows that human reactions can indicate when a robot fails \cite{2022stiber,spitale2024err,parreira2025neckface}, there is little research demonstrating how the progression of human reactions can be used to identify successive robot failures. Modeling these evolving features offers the potential to classify not only whether a robot has failed, but also the stage of failure within a sequence. This research explores the possibility of using machine learning to recognize successive robot errors from human reactions.

In prior work \cite{lbr}, we carried out a user study wherein 26 participants interacted with a robot which exhibited successive conversational failures. Participants' reactions were captured on videos and relevant behavioral features were extracted. We explore a range of machine learning strategies to detect successive robot error for a single user or across multiple participants. Model training on single participants allows systems to learn each individual's unique way of signaling robot errors, while training on multiple participants tests generalization to unseen participants. When training on single participants, the top performing model for detecting a robot error had an average accuracy of 93.5\% and the top performing model for classifying successive robot errors had an average accuracy of 84.1\% over the 26 participants. 


This research provides insight into the dynamics of successive interaction ruptures in HRI, by exploring how machine learning models can classify and generalize detection of repeated robot errors.

%% file: sections/methods.tex
\section{Methods}
\label{sec:methods}

\begin{figure}[ht]
\centering
\includegraphics[width=1\linewidth]{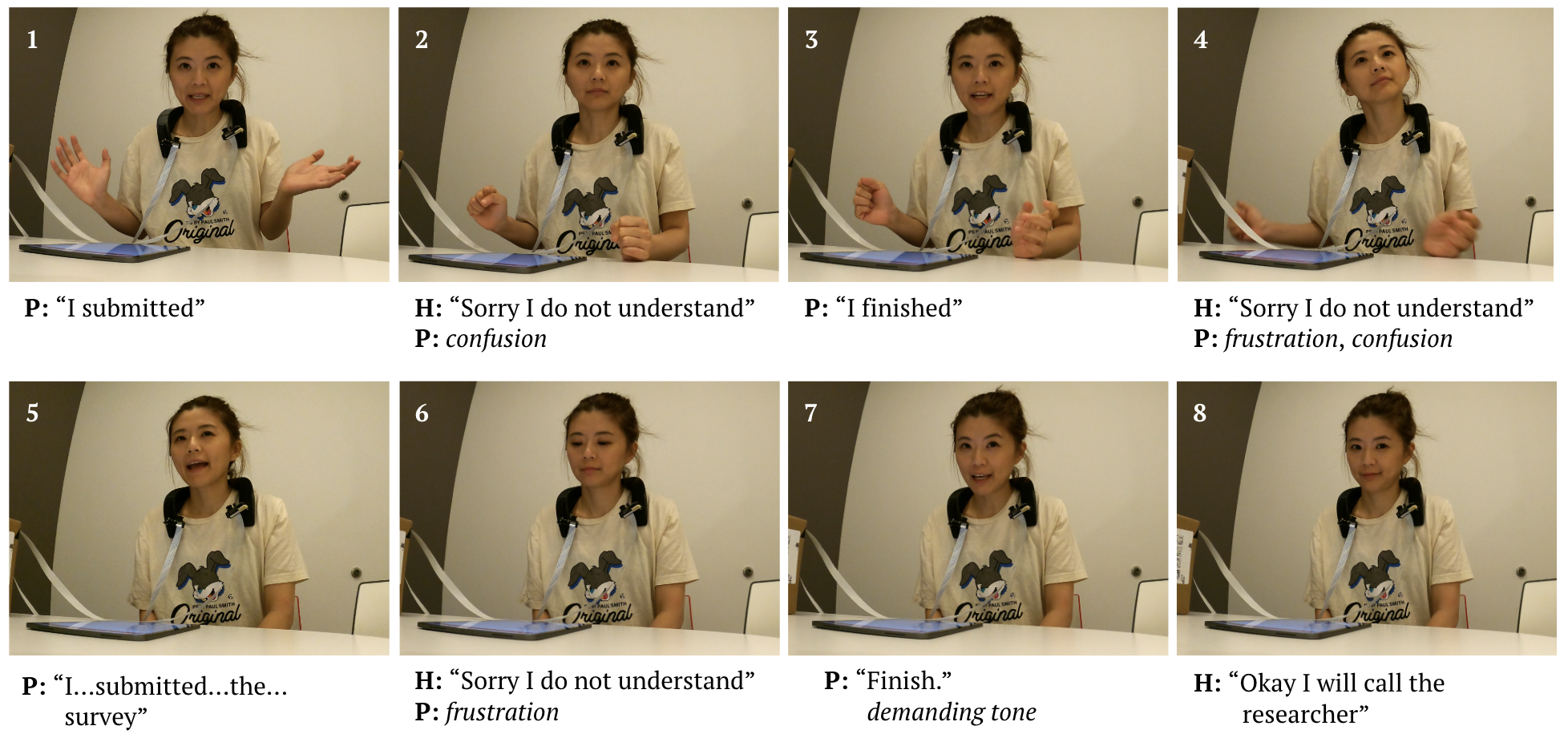}
\caption{A vignette of a participant's interaction with a robot (adapted from our earlier work \cite{lbr}).} \label{fig:vignette}
\end{figure}

 For each of the 26 participants, we extracted facial features (facial action units and gaze), body pose estimations (upper body keypoints), audio features, and text features, using OpenFace \cite{openface}, OpenPose \cite{openpose}, openSMILE \cite{opensmile}, and CLIP \cite{clip} and BERT \cite{bert}, respectively. \autoref{fig:vignette} shows an example of responses from a participant who interacted with a robot which failed 3 times successively. For each participant, video frames were annotated according to the occurrence of robot errors. In the Error Detection scheme, 
frames captured prior to any error were labeled as 0 (NoError), while frames captured after any error were labeled as 1 (AnyError). In the Successive Error Detection scheme, frames captured prior to any error were labeled as 0 (NoError) and frames captured after the first, second, and third error were labeled as 1 (Error1), 2 (Error2), and 3 (Error3), respectively. The extracted features were used as inputs to machine learning models for detecting successive robot errors which are represented by either binary or multiclass labels.

%% file: sections/models.tex
\section{Model Development}
\label{sec:models}

\begin{table*}[t]
\small
\centering
\caption{Model hyperparameters and test set performance. Metrics are shown in $M \pm SD$ for all participants for single-participant models ($n=26$). Each model was trained for 50 epochs.}
\begin{adjustbox}{center}
\begin{tabular}{c>{\centering\arraybackslash}p{40pt}>{\centering\arraybackslash}p{50pt}>{\centering\arraybackslash}p{50pt}>{\centering\arraybackslash}p{50pt}>{\centering\arraybackslash}p{50pt}>{\centering\arraybackslash}p{50pt}>{\centering\arraybackslash}p{50pt}>{\centering\arraybackslash}p{50pt}}
\hline
\textbf{Model} &  \textbf{Classification}&\textbf{Modalities}& \textbf{Representation} &\textbf{Fusion}& \textbf{Accuracy} & \textbf{Precision} & \textbf{Recall} & \textbf{F1} \\
\hline

 \multicolumn{8}{c}{\textit{Error Detection}} &\\ \hline
\rowcolor[HTML]{EFEFEF} 
LSTM &  Binary&facial& PCA&late& $0.935 \pm 0.032$ & $0.930 \pm 0.039$ & $0.923 \pm 0.041$ & $0.924 \pm 0.039$ \\
GRU  &  Binary&facial& PCA&intermediate& $0.922 \pm 0.039$ & $0.927 \pm 0.036$ & $0.919 \pm 0.040$ & $0.920 \pm 0.040$ \\ \hline

 \multicolumn{8}{c}{\textit{Multiple Error Detection}} &\\ \hline
\rowcolor[HTML]{EFEFEF} 
LSTM &  Multiclass&pose, audio&  Normalized&early& $0.782 \pm 0.071$ & $0.764 \pm 0.097$ & $0.700 \pm 0.113$ & $0.705 \pm 0.123$ \\
GRU  &  Multiclass&pose, facial, audio&  Normalized&late& $0.841 \pm 0.045$ & $0.824 \pm 0.059$ & $0.795 \pm 0.068$ & $0.800 \pm 0.064$ \\ \hline

 \multicolumn{8}{c}{\textit{First Error to Successive Errors Generalization}} &\\ \hline
\rowcolor[HTML]{EFEFEF} 
LSTM &  Binary&facial& PCA&late& $0.740 \pm 0.147$ & $0.759 \pm 0.151$ & $0.744 \pm 0.138$ & $0.726 \pm 0.163$ \\
GRU  &  Binary&facial& PCA&early& $0.723 \pm 0.129$ & $0.747 \pm 0.125$ & $0.727 \pm 0.126$ & $0.705 \pm 0.147$ \\ \hline


 \multicolumn{8}{c}{\textit{Successive Error Discrimination}} &\\ \hline
\rowcolor[HTML]{EFEFEF} 
LSTM &  Multiclass&facial, audio&  Normalized&early& $0.900 \pm 0.050$ & $0.899 \pm 0.056$ & $0.854 \pm 0.082$ & $0.858 \pm 0.081$ \\
GRU  &  Multiclass&facial, audio&  Normalized&early& $0.897 \pm 0.045$ & $0.880 \pm 0.058$ & $0.877 \pm 0.056$ & $0.873 \pm 0.061$ \\ \hline

\end{tabular}
\end{adjustbox}
\label{tab:model_results}
\end{table*}

We explored a range of machine learning strategies in order to evaluate how different data preprocessing, modality selection, model architectures, and fusion methods impact classification performance. Models were trained and evaluated using 26-fold cross-validation, where each participant served as one fold. For each fold, data was split into training, validation, and test sets according to the specified splitting strategies defined below. Training was run for 50 epochs per fold, and the final performance measure is reported as the average test accuracy across all folds.

\textbf{Data splitting:} We designed 4 data splitting methods. For each method, the data was randomly shuffled before partitioning. Across all experiments, we applied an 80/20 train–test split, with 10\% of the training set further reserved for validation.

\begin{enumerate}
    \item \textbf{Error Detection (binary):} Training and testing data included all binary labels (0 = NoError), 1 = AnyError).

    \item \textbf{Multiple Error Detection (multiclass):} Training and testing included samples from all multiclass labels (0 = NoError, 1 = Error1, 2 = Error2, 3 = Error3).
    
    \item \textbf{First Error to Successive Errors Generalization (binary):} Training included neutral frames (NoError) and reactions to the first error (Error1). 
    To balance the training set, neutral frames were randomly downsampled to equal the number of reactions to the first error. Testing data included the remaining neutral frames along with successive error reactions (Error2, Error3). 
    
    
    \item \textbf{Successive Error Discrimination (multiclass):} Training and testing included error reaction labels (Error1, Error2, Error3).
    
\end{enumerate}

\textbf{Feature representation:} We tested three different approaches to represent input features: (1) using raw non-normalized features, (2) applying normalization to ensure consistent feature scales, and (3) applying principal component analysis (PCA) to reduce dimensions.

\textbf{Modality combinations:} Facial, pose, audio, and text features were used as input to models. We performed modality ablation testing to find the best performing combination of modalities for each classification problem.

\textbf{Model architecture:} Two neural network architectures were explored: (1) Long Short-Term Memory (LSTM) networks to capture long-term dependencies in sequential data and (2) Gated Recurrent Units (GRUs) as a lighter-weight alternative.

\textbf{Fusion strategies:} We explored three fusion methods: (1) early fusion by concatenating features before model input, (2) intermediate fusion by processing modalities separately then merging, and (3) late fusion by training modalities separately and combining predictions.

Combinations of these strategies were systematically tested and evaluated to determine whether machine learning models can detect successive robot errors.


%% file: sections/results.tex
\section{Model Results}
\label{sec:results}

\autoref{tab:model_results} summarizes the top performing models along with configurations (modalities, feature representation, and fusion strategy) for each data splitting strategy and model type.

Error Detection achieved the highest overall performance (93.5\% accuracy), providing a strong baseline for detecting whether an error occurred. In contrast, First Error to Successive Error Generalization had reduced performance (74\% accuracy) which is likely caused by subtle variations in responses following successive errors. Successive Error Discrimination (90\% accuracy) achieved slightly higher accuracy than Multiple Error Detection (84.1\% accuracy), indicating that while both approaches capture the progression of human responses to successive errors, user reactions to repeated errors are more consistent and easier to distinguish than the broader range of error and no-error conditions.

%% file: sections/discussion.tex
\section{Discussion}
\label{sec:discussion}

Our experiments demonstrate that machine learning models can accurately detect robot errors based on human reactions, although performance varies depending on the modeling approach and type of error classification.


Because models were trained on individual participants, results highlight their abilities to learn each participant’s unique behavior. Binary strategies reliably detect errors, while multiclass and hybrid strategies capture successive errors and their progression.

%% file: sections/conclusion.tex
\section{Conclusion}
\label{sec:conclusion}

This study demonstrates that machine learning models can detect successive robot errors from human reactions and generalize across varied interactions within an individual. Since we did not evaluate our models on entirely unseen participants, our results may not directly measure generalization to new individuals. Future work should evaluate inter-participant generalization during training, which would more directly test robustness to unseen behaviors. Overall, our findings provide insight into how humans respond to repeated robot failures and suggest that HRI systems can leverage these signals to anticipate errors and improve performance.